\documentclass[journal]{IEEEtran}
\usepackage{amsmath,amssymb,amsfonts}
\usepackage{algorithm}
\usepackage[noend]{algpseudocode}
\usepackage{times}
\usepackage{epsfig}
\usepackage{graphicx}
\usepackage[caption=false]{subfig}
\usepackage{textcomp}
\usepackage{xcolor}
\def\BibTeX{{\rm B\kern-.05em{\sc i\kern-.025em b}\kern-.08em
    T\kern-.1667em\lower.7ex\hbox{E}\kern-.125emX}}
\begin{document}

\title{Recognizing Image Objects by Relational Analysis Using Heterogeneous Superpixels and Deep Convolutional Features}


\author{Alex~Yang, Charlie~T.~Veal, Derek~T.~Anderson,  Grant~J.~Scott%
\thanks{All authors are with the Department of Electrical Engineering and Computer Science, University of Missouri, Columbia, MO, 65211 USA.}%
\thanks{Corresponding authors: \texttt{zy5f9@mail.missouri.edu} and \texttt{GrantScott@missouri.edu}}}

\maketitle


\begin{abstract} 
  Superpixel-based methodologies have become increasingly popular in computer vision,
  especially when the computation is too expensive in time or memory to perform with a large number of pixels or features.
  However, rarely is superpixel segmentation examined within the context of deep convolutional neural network architectures.
  This paper presents a novel neural architecture that exploits the superpixel feature space.
 The visual feature space is organized using superpixels to provide the neural network with a substructure of the images.
  As the superpixels associate the visual feature space with parts of the objects in an image,
  the visual feature space is transformed into a structured vector representation per superpixel.
  It is shown that it is feasible to learn superpixel features using capsules and it is potentially beneficial to perform image analysis in such a structured manner.  
  This novel deep learning architecture is examined in the context of an image classification task, 
  highlighting explicit interpretability (explainability) of the network's decision making.
  The results are compared against a baseline deep neural model, 
  as well as among superpixel capsule networks with a variety of hyperparameter settings.
\end{abstract}



\section{Introduction}\label{sect:intro}

The landscape of computer vision has drastically changed since the resurgence of neural networks.
Neural networks, now popularized as deep learning models, have achieved state-of-the-art results in areas such as object recognition and localization \cite{ILSVRC, COCO}, semantic segmentation \cite{FULLYCONV, SEMSEG}, RGB-D \cite{DCNF} and multiple view data analytics, \textit{etc}. 
However, most deep learning methodologies usually require very high data processing bandwidth, and part of their overwhelming success can be attributed to the availability of large-scale image datasets \cite{ILSVRC, DL3D}. 
These deep learning models also create a large volume of features from their feature extraction process. 
The problem is that these models do not posses the following, regarding their generated feature space:
(i) an intelligent methodology to organize feature maps in more meaningful and comprehensible way;
(ii) a strong relationship between features and the spatial components of the observed samples. 
One of the forerunners \cite{DEFORM}, who is actively approaching these areas of concern, is the ideology of superpixel-enabled relational analysis. 

\begin{figure}
  \begin{center}
    \mbox{
      \includegraphics[width=0.9\columnwidth]{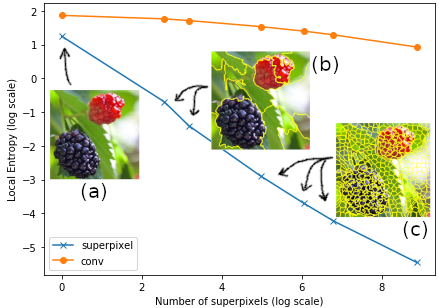}
    }
  \end{center}
  \caption{Superpixel local entropy: a quantity useful for classifying different usages of superpixels in machine learning models.
  The blue line (cross-points) indicates the mean of the superpixel entropy, while the orange line (round-dot-points) indicates a local entropy calculated similar to convolution where the local neighborhood is defined by a square window. For the convolving fashion local entropy, the window size used is that which has the same area as the average superpixel size, in order to compare against the superpixel local entropy.
 }
  \label{fig:sp-entropy}
\end{figure}

As Fig. \ref{fig:sp-entropy} shows, most superpixel models are confined to overly segmented superpixels associated with low local entropy and low information loss upon homogenization.
Contrary to popular belief, one can also see the value of superpixel usage in the slighly higher entropy range, which affords the opportunity to establish a relationship between features and their spatial properties.
The data points in Fig. \ref{fig:sp-entropy} are also listed in Tab. \ref{tab:sp-entropy}, which serve as an empirical evidence of the following observations:
(i) Compared to a convolutional operation, superpixels induce a mechanism more efficient in orders of magnitudes to reduce mean local entropy, which makes the superpixel feature space highly organized.
(ii) The \textit{de-facto} popular usage of superpixel as been dimensionality reduction through oversegmentation, whereas deep leaning models without superpixels may use techniques as drastic as global pooling which would be an extreme undersegmentation.
This means the homogeneity of superpixels alone shall not sufficiently determine the performance of a superpixel based model.
(iii) Empirically, the number of superpixels and the mean local entropy exhibit a strong anticorrelation (as demonstrated in Fig. \ref{fig:sp-entropy}) across a board range (from 1 to 10,000) of number of superpixels.
These are the observations that give rise to the model that is proposed herein, 
and the reasons that heterogeneous or high-entropy superpixel usage deserves to be classed as an entirely different type of model,
which is further discuss in Sect. \ref{sect:sp-feature}.


Prior researches \cite{DCNF} \cite{SPSEG} in the field of deep learning have used superpixels mainly to reduce the computational intensity.
Superpixel pooling is one of the primary recognized techniques of dimensionality reduction for deep convolutional networks.
This method allows perceptually homogeneous regions of pixels to be grouped together and represented more concisely \cite{SLIC},
and then aggregated according to such regions. 
As reported in the literature, such as \cite{DCNF, SPCNN}, it is feasible to adapt superpixel features to many deep convolutional neural network (DCNN) architectures to extract features from convolutional maps,
making a potentially very expensive computation feasible.
%
%
More interestingly, however, in some literature it is reported that not only have superpixels made the computation more efficient, but there has been model performance benefits as well.
For example, in the work of Mostajabi, \textit{et al.} \cite{SPSEMSEG} superpixels were used to exploit statistical structure in the image and in the label space and to avoid complex and expensive inference.
More recently, in the work of Wu, \textit{et al.} \cite{SPCNN} superpixel based analysis is incorporated to perform highly efficient cloud image segmentation. 
The local consistency of the segmentation was preserved and faster DCNN learning was reported.

Consequently, with respect to the above, superpixels can also provide an alternative organization of the convolutional features. 
Herein, superpixels are utilized to reorganize the feature space by associating convolutional features to discrete parts of objects within images. 
Afterwards, relational analysis is explored, in the form of capsule routing, to fuse structurally organized convolutional features. 
Relational analysis in this sense, will prioritize the spatial relationship between the features while combining them into a set of model predictions \cite{CAPS}.

\begin{figure}
  \begin{center}
    \mbox{
      \includegraphics[width=0.9\columnwidth]{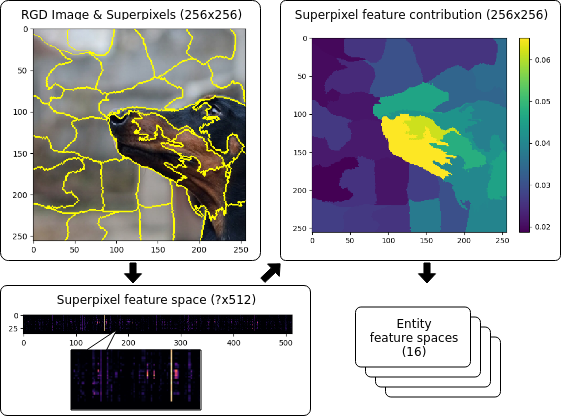}
    }
  \end{center}
  \caption{Feature space reorganization using superpixel:
  On the top right corner is a conceptual illustration, not actual step of computation, of superpixel feature contribution.
  The input image is first processed by a VGG-16 network before the superpixel features are extracted.
  The superpixel feature is represented in a matrix, for example $36\times512$ where each of the 36 superpixels would have 512 feature components.
  The superpixel features are used for image object relational analysis with several isolated entity feature spaces.
  As the superpixel segmentation still retains the spatial relation between the superpixel features, the superpixel feature contribution can be visually reconstructed as shown in the top right corner. 
  It can be see the superpixels that made up the dog are producing a higher level of contribution to the determination of the class dog, as expected.
 }
  \label{fig:sp-feature-space}
\end{figure}


The challenge with processing superpixel features and learning a model that makes inferences based on superpixels,
is to fully exploit the superpixel features with some capable mathematical machinery; 
as opposed to performing naive aggregations too early and losing a significant amount of information.
To this end, capsules \cite{CAPS} are experimented with to create a model that analyzes visual images by putting together pieces of superpixels and processing the superpixel vector features.
As each superpixel segmentation of the image is completely different, there is no specific role assigned to the neurons handling the superpixels and this makes the superpixel feature space difficult to learn especially with high dimensionality.

Related state-of-the-art research actively addressing the connection between feature space and feature spatial properties also includes Deformable ConverNets \cite{DEFORM},
which has allowed the region of interest to adapt to the geometric variations of objects.
Compared to the superpixel approach this geometrical adaptation is not limited to perceptually homogeneous regions and it would also relax the disjoint constraint.
The similarity shared with the superpixel approach, including the proposed method, is the acknowledgment  that introducing irregular regions that account for geometrical cues is considerably
enhancing of the modeling capability of a deep convolutional neural network.

%
The strength of the proposed model primarily lies in the heterogeneous superpixel usage, which has facilitated image object relational analysis,
thanks to its capability to associate convolutional features to their spatial locality, and in turn, parts of objects in the image.
The usage of heterogeneous superpixel does not follow the popular homogeneity assumption, but have demonstrated a preliminary success in the effort of image object recognition based on relational analysis,
while most DCNNs that do not use superpixel either aggregate across the entire image or process at the pixel level, as illustrated in Fig. \ref{fig:sp-entropy} (a),
and  most current superpixel based DCNNs or PGMs rely on oversegmentation, as illustrated in Fig. \ref{fig:sp-entropy} (c).
Although the relational analysis appears to be similar to segmentation, it is worth pointing out that it is not feasible to compare with the state-of-art segmentation DCNN such as U-Net \cite{UNET} or various other segmentation networks \cite{pal2018capsdemm} \cite{lalonde2018capsules} that rely on it, because those types of neural networks are using segmentation labels for training while the proposed model has none, instead implicitly learning through many occurrences of the image objects given non-localized class labels only.
Therefore, the model would be more accurately described as multiple-instance learning from a segmentation model point of view.
By computing the superpixel-wise feature contribution to the class vector, visualizations are generated that represent such part-whole relations and make it more feasible to segment the object of interest as a result.
In addition, the image substructure provided by superpixels can be beneficial to other related problem domains such as localization, segmentation, and data fusion.

The remainder of this paper is structured as follows.
Sect. \ref{sect:sp-feature} details the mathematics and theoretical considerations of the superpixel feature extraction, followed by superpixel feature aggregation in Sect. \ref{sect:sp-agg}.
Sect. \ref{sect:eval} details the evaluation framework, followed by a discussion of the dataset and novel neural architecture.
Then, experimental results are discussed in Sect. \ref{sect:results}.
Finally, Sect. \ref{sect:conclusion} provides concluding remarks and discusses potential future investigations.

\section{Superpixel features}\label{sect:sp-feature}

Superpixel segmentation is conventionally a way of grouping pixels into perceptually homogeneous regions,
which provides an efficient representation of image data.
However, superpixels have been used in a way that does not comply with the popular assumption; 
specifically, they are extended to the concept of homogeneous regions, a scope that is beyond perceptual or visual.
Herein, relational analysis is facilitated by allowing heterogeneous superpixel;
and allowing the superpixel local entropy to be an order of magnitude higher than typical superpixel usage, as shown in Tab. \ref{tab:sp-entropy}.
Relative to the traditional usage, this can be referred to as \textit{heterogeneous superpixelation}.
The SLIC \cite{SLIC} superpixelation algorithm is used, 
which is a $k$-means clustering based algorithm to efficiently generate an approximately fixed number $k$ of superpixels as a segmentation.
The resulting superpixels of this algorithm are particularly well suited for neural networks when the neural network has preallocated tensor shapes.
Additionally, it has linear time complexity $O(N)$ \cite{SLIC} with respect to the number of source pixels $N$;
and it can be implemented using parallelism \cite{gSLIC} to drastically accelerate the segmentation.

\begin{table}
\caption{Symbols and Notation}
\label{tab:notation}
\centering
\begin{tabular}{rp{2.5in}}
\hline
\hline
$V$ & Denotes the entire pixel set of an image. \\
$S$ & Number of superpixels. \\
$T$ & Equivalent convolutional window size, used in Tab. \ref{tab:sp-entropy}. \\
$Q$ & class vector size, used in Tab. \ref{tab:train}. \\
$x$ & An image is defined based on a mapping from a pixel set to the real numbers: $V \to \mathbb{R}$. \\
$R_s$ & Superpixels are represented by pixel sets as well, which can be any arbitrary regions, identified by a subscript. \\
$\textbf{u}_j$ & Feature vectors, where the subscript is to identify these vector themselves as opposed to their components.\\
$(\textbf{u}_j)_k$ & the $k$-th component of vector $\textbf{u}_j.$\\
\hline
$\mathbb{R}$ & The set of real numbers. \\
$\textbf{u}$  & A vector with component form: $\textbf{u} \in \mathbb{R}^{n}. $\\
$\textbf{W}$  & A matrix with component form: $\textbf{u} \in \mathbb{R}^{m\times n}. $\\
$\mathcal{P}(\cdot)$ & The power set operator, which generates all subsets.\\
$\left| \cdot\right|$ & The cardinality of a set, also known as the number of elements for finite sets.\\
$\| \cdot \|$ & The magnitude or the $L^2$-norm of a vector.\\
$ \circ $ & function composition: $(f\circ g)(x) =f(g(x))$ .\\
\hline
\end{tabular}
\end{table}

\subsection{Superpixel Homogeneity}

The superpixel segmentation will determine the extracted features and affect the structure of the neural network,
as such, the parameters used to generate superpixels become new hyperparameters for the neural network.
The SLIC superpixel algorithm operates by first converting the image's color into CIE L*a*b color space,
then clustering based in 5-D feature space composed of the concatenation of  CIE L*a*b and the 2-D pixel coordinate (spatial) system.
The color proximity and spatial proximity are weighted in the form of a compactness factor $m$.
The distance measure $D$ used in the SLIC algorithm is shown in Eq. (\ref{eq:slic-distance}),
where $s$ is the sampling interval, which controls the grid geometry upon initialization; $d_c$ is color proximity; $d_s$ is spatial proximity \cite{SLIC};
\begin{equation}
  \label{eq:slic-distance}
D(p_1, p_2) = \sqrt{d_c^2(p_1, p_2) + \left(\frac{d_s(p_1, p_2)}{s}\right)^2 m^2} ~ .
\end{equation}
\nobreak

The number of superpixels, $k$, will control the granularity of the superpixel segmentation, 
and the compactness factor and smoothing factor will control the superpixel shape irregularity that is achievable.
As a result, these superpixel hyperparameters allow us to effectively extract and represent convolutional features
and efficiently adapt computational resources to process them.

As mentioned, the model has allowed a relatively high superpixel local entropy as an admissible segmentation.
The mean local entropy is measured by creating a 256-bin hue histogram with respect to each superpixel region.
From Fig. \ref{fig:sp-entropy}, it is evident that in the context of superpixel segmentation, the mean local entropy
exhibits a strong anticorrelation with respect to number of superpixels.
Superpixels are viewed as disjoint pixel sets $R_i$ and the entire image as the pixel set $V$ and $H(\cdot)$ as the entropy operator over some pixel set. 
Based on the properties of entropy, Eq. \ref{eq:partitioned-entropy-general-bound}

\begin{equation}
  \label{eq:partitioned-entropy-general-bound}
  0 \leq \sum_{i=1}^S \frac{|R_i|}{|V|} H(R_i) \leq H(V)
\end{equation}

\noindent is the general bound for superpixel local entropy, which means

\begin{equation}
  \label{eq:partitioned-entropy-general-bound2}
  \exists M \in [0,1]:  \sum_{i=1}^S \frac{|R_i|}{|V|} H(R_i) = M \cdot H(V) ~ .
\end{equation}

\noindent Moreover, empirically it can be found that $M \approx S^{-1} \in [0,1]$ is the solution according to Tab. \ref{tab:sp-entropy} across a broad range of number of superpixels evaluated.
Therefore, the multiplication factor $M$ has shown the effect of reducing entropy of superpixel feature organization; 
and it is strongly dependent on the number of superpixels used with a given image resolution.

\subsection{Superpixel Feature Extraction}\label{sect:sp-extraction}

\begin{table}
	\begin{center}
	\caption{Relation between superpixel size and local entropy}
    \label{tab:sp-entropy}
	\begin{tabular}{rr|rr|rr}
		\hline
		\hline
		\multicolumn{2}{c}{Superpixel Size} & \multicolumn{2}{|c|}{Compression (\%)} & \multicolumn{2}{c}{Entropy} \\
	    S & T & Input & Conv. & Superpixel & Conv. \\
		\hline
		1 & 256 & 0.0015 & 1.3888 & - & 6.4720  \\
		\textbf{13} & 71 & 0.0198 & 18.0555 & \textbf{0.4993} & 5.8326  \\
		\textbf{24} & 52 & 0.0366 & 33.3333 & \textbf{0.2457} & 5.5490 \\
		145 & 21 & 0.2212 & 201.3888 & 0.0552 & 4.6355 \\
		425 & 12 & 0.6484 & 590.2777 & 0.0246 & 4.0566 \\
		894 & 8 & 1.3641 & 1241.667 & 0.0145 & 3.6366  \\
		7185 & 3 & 10.9634 & 9979.167  & 0.00426 & 2.5295 \\
		\hline
		\hline
		\multicolumn{6}{l}{\footnotesize $S$ is the number of superpixels, equivalent to stating superpixel size.}\\
		\multicolumn{6}{l}{\footnotesize $T$ is the equivalent convolutional window size to compare local entropy.}\\
		\multicolumn{6}{l}{\footnotesize The compression ratio is w.r.t. input image size and $8\times 8$ conv.}\\
		\multicolumn{6}{l}{\footnotesize feature maps respectively in terms of number of elements.}\\
		\multicolumn{6}{l}{\footnotesize The ``Conv.'' entropy is the conventional sense of local entropy calculated}\\
		\multicolumn{6}{l}{\footnotesize using a sliding window, as its operator works like convolution, renamed it}\\
		\multicolumn{6}{l}{\footnotesize to distinguish from superpixel local entropy.}\\
	\end{tabular}
	\end{center}
\end{table}

As discussed, a superpixel segmentation algorithm provides a partitioning of a 2-D image with irregular region shapes.
Thus a superpixel-pooled image can be represented as a series of superpixel intensity values.
The resulting superpixel-pooled image is usually stored in a 1-D data structure with multiple channels.
A 2-D spatial representation can be obtained by filling superpixel intensity values back into the segmentation.

Consider a pixel set $V$ that contains all the pixels from an image $x$ is partitioned into many (typically) disjoint regions (superpixels).
Each disjoint region is assigned an aggregated value over the region $R$;
and therefore, superpixel representation is obtained from the feature space as Eq. (\ref{eq:sp-def}). 

\begin{equation}
    \label{eq:sp-def}
	y_j=\frac { \sum _{p \in R_j} x(p) }{ \left| R_j\right| } ~.
\end{equation}

Given that $R$ is chosen to respect part-whole relationships in the image, 
features are reorganized to respect substructures within the image, facilitating structure awareness.
This reorganization retains the association between vectors from the feature space and their locality within the image.

Another technical detail to be addressed is that the resolution of the convolutional feature maps and the superpixel segmentation are likely not to match.
The superpixel segmentation is computed based on the 2-D visual image,
but superpixel feature extraction will be applied to features extracted by a deep convolutional network.
This difference in resolutions is addressed by associating each pixel in the convolutional maps to a tile of pixels from the segmentation.

The value of any pixel in an image ($x: V \to \mathbb{R} $) up-scaled, as opposed to up-sample because there is no up-sampling algorithm involved here, by an integer factor (to $x': V \to \mathbb{R} = x \circ \tau$) can be found directly from $x$. 
Eq. (\ref{eq:sp-substitution}) provides a rule for substitution, where $\tau: p \in V \mapsto ( \lfloor p_0/t \rfloor, \lfloor p_1/t \rfloor)$ is the coordinate transformation, $t$ being the tile size, from the up-scale image to the original image and $T(i,j) = T_{ij} \in \mathcal{P}(V)$ represents the tiles that pixels from $x$ are up-scaled into,

\begin{equation}
   \label{eq:sp-substitution2}
  \begin{aligned}
	\forall p\in T_{ij}&: (x' \circ T)(i, j) = x'(T_{ij}) = x'(p) \\
	     &= (x \circ \tau) (p) = x(\tau(p)) = x(i,j) ~.
	\end{aligned}
\end{equation}

\noindent in short,

\begin{equation}
  \label{eq:sp-substitution}
  x = x' \circ T ~ .
\end{equation}

\noindent This association will introduce another way of partitioning the 2-D image, where it is necessary to further generalize superpixel feature extraction to partitions of an image.
From now on, the up-scaled image is ignored because Eq. (\ref{eq:sp-substitution}) has made it feasible to directly associate pixels across different feature map resolutions.
Such direct association is pointed out in \cite{DCNF} as well, but it is demonstrated that it is accurate with Eq. \ref{eq:sp-substitution2}.

On arbitrary partitions of $V = \cup V_i$, where a single intensity value $x(V_i)$ can be assigned to each partition,
each superpixel will relate to many partitions and the aggregation over repeating pixels becomes multiplication,

\begin{equation}
  \label{eq:sp-partition}
	y_{j}=\frac { \sum _i  {\left| R_j \cap V_i \right|x(V_i)} }{ \left| R_j\right| } ~ .
\end{equation}







The superpixel feature extraction has no parameters to be trained, only the gradient between the input and output needs taken into account during training.

\noindent 
Therefore, superpixel feature extraction may be utilized to efficiently aggregate homogeneous pixel regions,
while avoiding up-scaling convolutional maps.
It flattens the feature maps to 1-D, where in many cases, it is easier to further process,
without necessarily losing 2-D spatial properties, as a 2-D representation is always available as well.
The superpixel features are 1-D vectors, 
and the superpixel feature map has a $(n, s, k)$ shape, where $n$ is the batch size; 
$s$ is the fixed number of superpixels; and $k$ is the number of feature channels.

\begin{figure}
  \begin{center}
    \mbox{
      \includegraphics[width=\columnwidth]{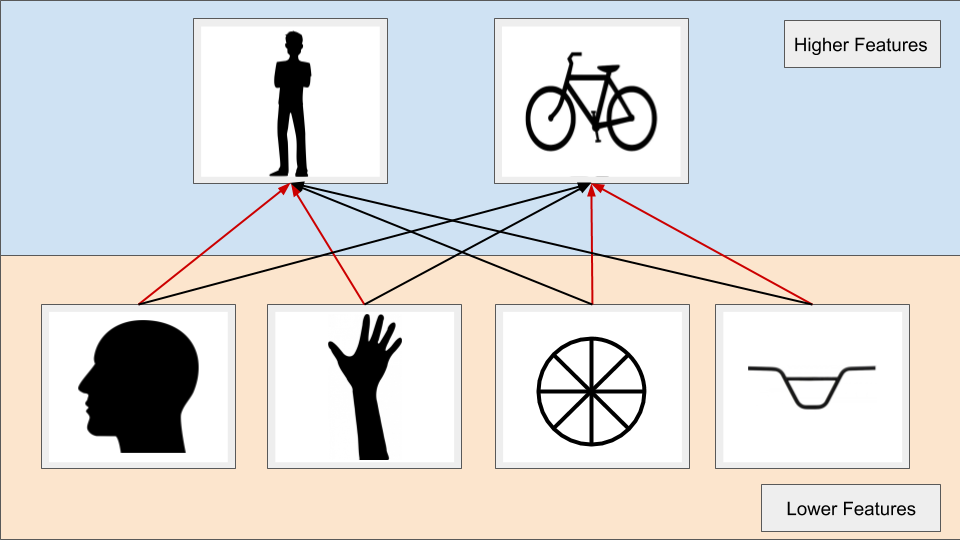}
    }
  \end{center}
  \caption{Visualization of relational analysis. This figure is broken up into two sections: (i) lower level superpixel features (ii) high level class features.
  Ideally, capsule routing exhaustively maps lower level features (the parts) to the appropriate higher level  features (the whole) through a chosen similarity metric. In the model, both the lower level features and the higher level features are represented by capsules. }
  \label{fig:routing}
\end{figure}

\section{Relational Analysis}\label{sect:sp-agg}

Deep learning has many situations that require an intelligent feature aggregation. 
Naturally, the multi-layer-perceptron has been the tool of choice, given its inherit ability to invoke universal function approximation.
However, 
care must be taken to choosing the aggregation because some may prove detrimental to the work done by superpixel extraction.
Sabour \textit{et al.} \cite{CAPS} put forward an ideology known as capsules to improve a model's knowledge of how its observed features relate to one another in a hierarchical manner.
This means that capsules try to mathematically model the relationship between low level features--the parts or children--and higher level features--the entity or parents. 
To accomplish this, capsules undergo a process known as dynamic routing. 
In Fig. \ref{fig:routing}, 
one can see that the dynamic routing algorithm allows capsules of lower level features
to send their input to capsules of higher level features;
More information about capsules and routing can be found in \cite{CAPS} and \cite{EMCAPS}.

Now accounting for $k$ channels in the superpixel feature map,
superpixel features form a matrix $[y_j^k]$.
The superpixel features are treated as if they were lower level capsule features $(\textbf{u}_j)_k = y_j^k $
and use dynamic routing to interface with the capsule layer that follows,

\begin{equation}
  \label{eq:spcaps}
  \textbf{v}_j = \phi \left( \sum_i c_{ij} {\textbf{W}_{ij} \textbf{u}_i} \right) ~ ,
\end{equation}

\noindent or more explicitly,

\begin{equation}
  \label{eq:spcaps2}
  v_j^{k_1} = \phi_{k_1'}^{k_1} \left( \sum_{i,k_0} c_{ij} {W_{ijk_{0}}^{k_1'} y_i^{k_0}} \right) ~ ,
\end{equation}

\noindent where $\textbf{v}_j$ are capsule outputs per class, which are entity instantiation vectors \cite{CAPS}; and
$\phi(\cdot)$ is a function that monotonically affects only the magnitude of the $k_1$-vector, whose exact form (see squash function \cite{CAPS}) is irrelevant here.
In Eq.~(\ref{eq:spcaps2}), $k_0$ and $k_1$ are the numbers of the input and output channels respectively.

The probability of an object that exists in the image is proportional to the magnitude of the instantiation vector,
which has a number of components that capture, in theory, various kinds of characteristics about the object in question \cite{CAPS}.
However, it should also be of interest to computer the individual superpixel features' contribution to the entity instantiation vector magnitude
because they can be projected back onto the superpixel segmentation to obtain a 2-D segmentation of the image,

\begin{equation}
  \label{eq:spcaps-segmentations}
  z_{ij} = \frac{ \left( \textbf{W}_{ij} \textbf{u}_i \right)^T \textbf{v}_j }{\| \textbf{v}_j \|} ~ ,
\end{equation}

\begin{figure}
  \begin{center}
    \subfloat[]{%
      \label{fig:spcaps-3col-in}
      \includegraphics[width=0.3\columnwidth]{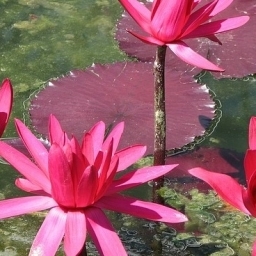}
    }
    \subfloat[]{%
      \label{fig:spcaps-3col-10}
      \includegraphics[width=0.3\columnwidth]{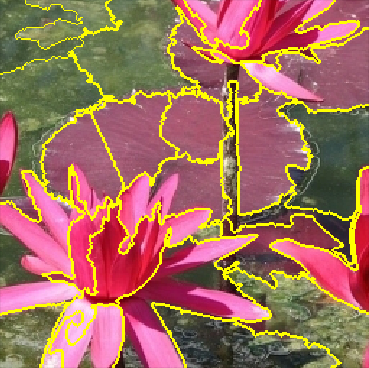}
    }
    \subfloat[]{%
      \label{fig:spcaps-3col-16}
      \includegraphics[width=0.3\columnwidth]{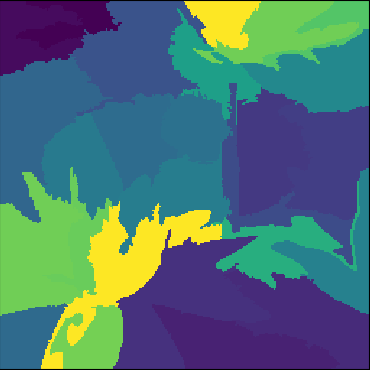}
    }
  \end{center}
  \caption{An example of entity segmentation using superpixel feature contribution (see Eq. (\ref{eq:spcaps-segmentations})): (a) sample input image; (b) superpixel segmentation; (c) entity contribution to recognition. }
  \label{fig:spcaps-segmentations-3col}
\end{figure}

Eq. (\ref{eq:spcaps-segmentations}) describes the computation of internal superpixel segmemtations, as a scalar projection,
with superpixels labeled by $i$ and the target entities labeled by $j$.
Fig. \ref{fig:spcaps-segmentations-3col} has shown a visual example of the segmentations.

\section{Proposed Network Architecture}\label{sect:eval}

\begin{figure}
  \begin{center}
    \mbox{
      \includegraphics[width=\columnwidth]{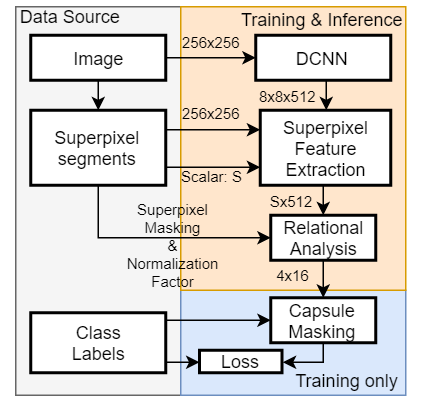}
    }
  \end{center}
  \caption{Proposed Network architecture: 
  Masking is used to selectively train these capsules.
  Only the class labels are supplied and the trained model is designed to perform entity segmentation and classification, through the part-whole relation modeling without pixel level or superpixel level segmentation labels.}
  \label{fig:short}
\end{figure}

In order to evaluate the effectiveness of image object recoginition based on superpixel relational analysis, comparisons are set up  between a DCNN model with and without superpixel feature extraction.
The superpixel feature extraction fulfills reorganizing the feature space in order to facilitate the part-whole relational analysis while maintaining the spatial association.
Therefore, one can back-trace the contribution brought by an individual superpixel once the model is trained and an entity segmentation is obtained.
The evaluated models share a common VGG-16 DCNN to generate the base convolutional feature maps,
but the superpixel size is varied across different models, 
which was the hyperparameter that has had the greatest impact on the model accuracy and speed of convergence during training.
As one can see that the superpixel feature space reorganization is based on visual features, various kinds of DCNNs can be utilized as visual feature extractors.
In this paper, VGG-16 \cite{VGG} was chosen for to two main reasons:
(1) to not overcomplicate the exploration of the ideology of superpixel feature based networks, 
direct experimentation on very large scale dataset at this stage to facilitate better inspection of the data processing; and 
(2) if the new feature spaces are a better internal representation for object recognition, 
then the network should not require as a deep neural network architecture as most popular deep learners \cite{CAPS}.
VGG16 was able to achieve 92.3\% top-5 accuracy on 1000-class ImageNet dataset.
VGG16 was chosen as a good starting point for a basic visual feature extractor because it has achieved high accuracy while involving relatively fewer parameters.
The authors of \cite{VGG} have released their best-performing models to facilitate further research.

The pretrained VGG16 network weights are used as initial state of training.
This technique is known as transfer learning, which allows models to learn faster by
leveraging existing weights from a related problem domain,
and transfer this knowledge gained in solving the source problem to solving the problem of interest.

As the VGG-16 DCNN feature extractor generates $8\times 8$ visual features with 512 channels, the superpixel feature extraction maps the $8\times 8$ visual features to the superpixels according to Eq. (\ref{eq:sp-def}).
The superpixel feature extraction requires convolutional features and a superpixel segmentation,
 and the resulting superpixel feature map will have the same amount of channels as the convolutional feature maps.
 As a result, the superpixel feature map is a 2-D data structure $(S \times k)$ per image sample, but neither axis is a spatial dimension of the image;
 instead, they are superpixels $S$ and channels $k$
 because the spatial axes of the visual features are merged into one, and can be restored given the superpixel segmentation.

Lastly, the relational analysis layer is implemented as 4 capsules with classes encoded as a 16-vector, where the probability of the entity existing in the image is represented by the magnitude of the class vector\cite{CAPS}.
To carry out the relational analysis, the  $S \times k$ superpixel feature map is interpreted as $S$ low-level capsules, which encode low-level features as a $k$-vector.

\subsection{Training}

Using stochastic gradient descent with a learning rate of $2\times10^{-5}$, various Superpixel Capsule Network models were trained with different superpixel sizes for 120 epochs.
All the models are initialized with pre-trained VGG-16 weights from ImageNet \cite{VGG}.
The 120-epoch limit does not necessarily drive the models to a learning saturation point, 
but the goal is to set up a comparable training setting between different models, such that they all converge in a relatively stable manner.
The size of the superpixels is important because it defines the scale of image parts whose features will be examined by capsules.
As the smoothing factor and compactness is relatively less important among the 4 experiments, $\sigma=0$ and $m=0.1$ are used throughout the experiments.
The superpixel sizes are constrained by specifying an approximate number of superpixels to be expected out of the superpixel segmentation algorithm.



\section{Experiments and Results}\label{sect:results}

The Superpixel Capsule Network is evaluated with a variety of hyperparameter settings and have also compared to a standard VGG-16 network for reference.
The model is based on the part-whole relationship concept with superpixel-based image analysis, 
so the key hyperparameters that are focused on are the size of the superpixels and the class vector dimension.
Table \ref{tab:train} summarizes the training performance of these variations of the model.
These new models have 88\% reduction in the total trainable parameters compared to VGG-16,
and can achieve 89\% accuracy with very good generalization ability on the evaluation dataset.
Additionally, it is shown that the internal representation of part-whole relationship in these models can be visually confirmed and is potentially very valuable for semantic segmentation and object localization purposes where pixel-level or superpixel-level segmentation label are not available.
Furthermore, exposure of the network's decision making in superpixel feature aggregation moves towards explainability of the classification decisions and valuable insights on the models learning, as the explicit interpretation of the contribution of superpixels to classification is available.

Due to the significant dimensionality reduction effect of superpixel pooling,
datasets with low resolution, such as CIFAR or MNIST, will not be feasible for this evaluation.
Therefore, the dataset of choice is \textit{Linnaeus 5} dataset \cite{L5},
which was created to evaluate object classification techniques.
This relatively small-scale dataset is used to better understand 
how the network is handling the transformation between
superpixel and entity feature spaces and the characteristics of convergence during training.
It has 4 classes: berry, bird, dog and flower and other unclassified images as adversarial examples.
The greatest resolution ($256 \times 256$) of the published versions is used.
Each of the 5 categories contains 1200 training images and 400 test images.
A 10-fold cross-validation methodology is used to compare the models.

\begin{table}[t]
	\begin{center}
	\caption{Training results of Superpixel Capsule Network using 88\% fewer parameters than VGG-16}
	\begin{tabular}{rr|rr|rr}
		\hline
		\hline
		\multicolumn{2}{c}{Model variations} & \multicolumn{2}{|c|}{Training} & \multicolumn{2}{c}{Validation} \\

	    S & Q & Loss & Accuracy & Loss & Accuracy \\
		\hline
		(VGG-16) & N/A & 0 & 1 & 0.0635 & 0.96  \\
		\hline
		10 & 16 & 0.1983 & 0.84 & 0.2004 & 0.83  \\
		16 & 16 & 0.1742 & 0.87 & 0.1799 & 0.87 \\
		25 & 16 & 0.1590 & 0.89 & 0.1718 & 0.88 \\
		36 & 16 & 0.1607 & 0.89 & 0.1697 & 0.88 \\
		50 & 16 & 0.1622 & 0.89 & 0.1701 & \textbf{0.89} \\
		100 & 16 & 0.1674 & \textbf{0.90} & 0.1708 & \textbf{0.89} \\
		200 & 16 & 0.1668 & 0.89 & 0.1844 & 0.85 \\
		\hline
		10 & 64 & 0.2551 & \textbf{0.81} & 0.2574 & \textbf{0.81} \\
		16 & 64 & 0.2570 & 0.79 & 0.2609 & 0.80 \\
		25 & 64 & 0.2578 & 0.79 & 0.2589 & 0.78 \\
		36 & 64 & 0.2622 & 0.76 & 0.2659 & 0.74 \\
		50 & 64 & 0.2654 & 0.76 & 0.2731 & 0.73 \\
		100 & 64 & 0.2642 & 0.68 & 0.2666 & 0.63 \\
		200 & 64 & 0.2587 & 0.6794 & 0.2611 & 0.61 \\
		\hline
		\hline
	\end{tabular}
	\label{tab:train}
	\end{center}
\end{table}

\subsection{Superpixel Homogeneity Versus Receptive Field}

\begin{figure}
  \begin{center}
    \subfloat[]{%
      \label{fig:compare_acc_16}
      \includegraphics[width=0.9\columnwidth]{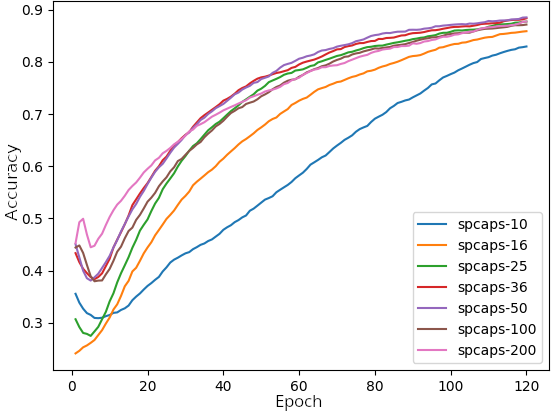}
      
    }
    
    \subfloat[]{%
      \label{fig:compare_loss_16}
      \includegraphics[width=0.9\columnwidth]{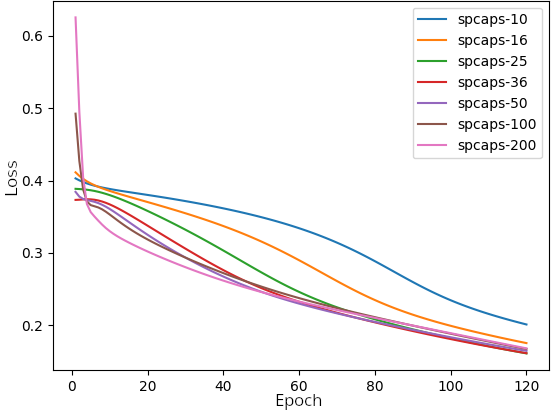}
    }
  \end{center}
  \caption{Comparison of training accuracy and loss between different superpixel sizes, which are indirectly constrained by specifying the approximate number of superpixels per image.
  Using 16-component vectors to represent entities, the general trend appears to be that the accuracy increases as smaller superpixel sizes are utilized.
  It can also be seen that the superpixel size affects convergence speed especially with smaller superpixel sizes.
  The relation between superpixel size and prediction accuracy is not always monotonic, as training metrics on the 64-component class vector experiments, in Table \ref{tab:train} has turned out to be a counterexample.}
  \label{fig:compare_16}
\end{figure}

\begin{figure*}
  \begin{center}
    \subfloat[]{%
      \label{fig:spcaps-in.1}
      \includegraphics[width=0.5\columnwidth]{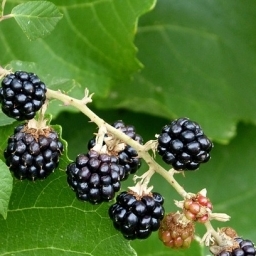} 
    }
    \subfloat[]{%
      \label{fig:spcaps-16.1}
      \includegraphics[width=0.5\columnwidth]{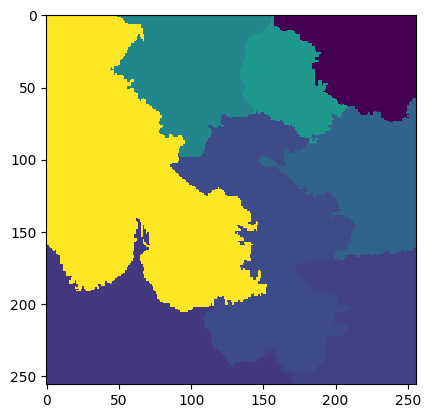}
    }
    \subfloat[]{%
      \label{fig:spcaps-36.1}
      \includegraphics[width=0.5\columnwidth]{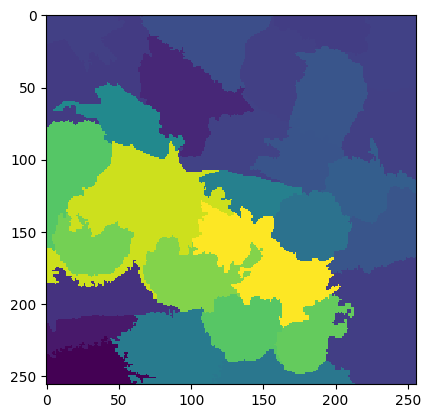}
    }
    \subfloat[]{%
      \label{fig:spcaps-100.1}
      \includegraphics[width=0.5\columnwidth]{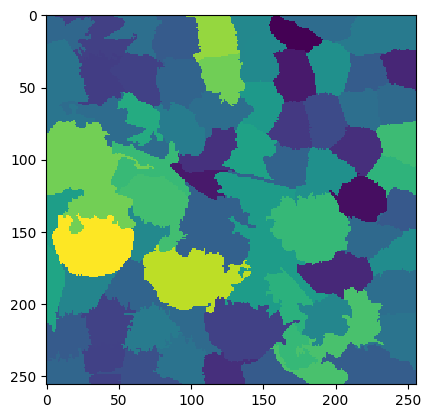}
    }
    
    \subfloat[]{%
      \label{fig:spcaps-in.2}
      \includegraphics[width=0.5\columnwidth]{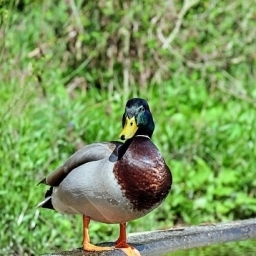}
    }
    \subfloat[]{%
      \label{fig:spcaps-16.2}
      \includegraphics[width=0.5\columnwidth]{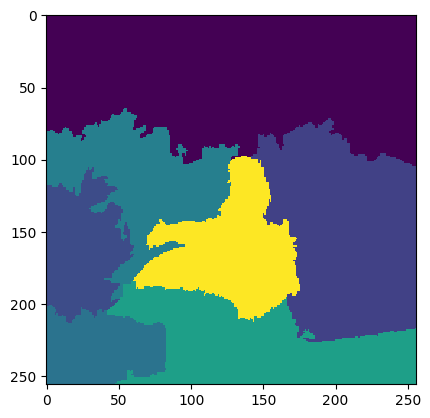}
    }
    \subfloat[]{%
      \label{fig:spcaps-36.2}
      \includegraphics[width=0.5\columnwidth]{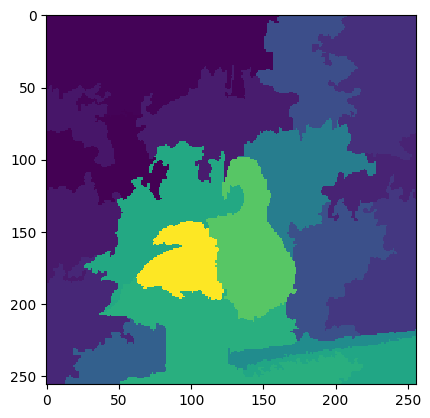}
    }
    \subfloat[]{%
      \label{fig:spcaps-100.2}
      \includegraphics[width=0.5\columnwidth]{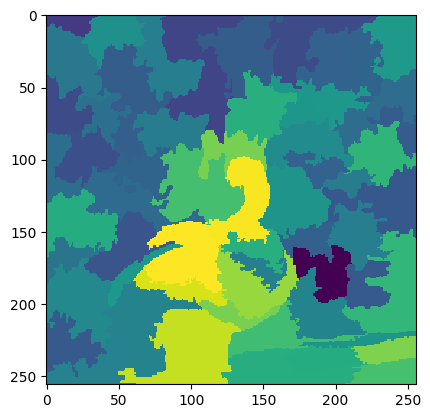}
    }
    
    \subfloat[]{%
      \label{fig:spcaps-in.3}
      \includegraphics[width=0.5\columnwidth]{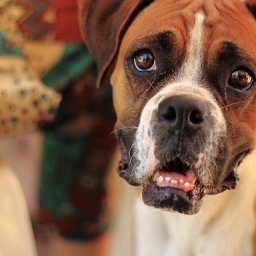}
    }
    \subfloat[]{%
      \label{fig:spcaps-16.3}
      \includegraphics[width=0.5\columnwidth]{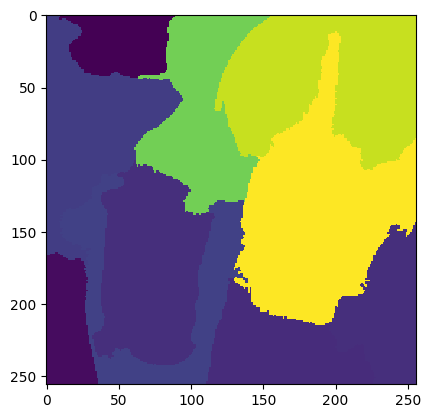}
    }
    \subfloat[]{%
      \label{fig:spcaps-36.3}
      \includegraphics[width=0.5\columnwidth]{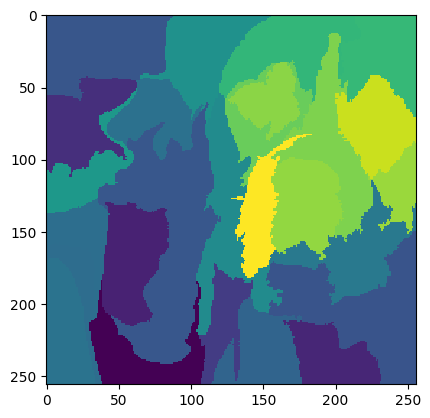}
    }
    \subfloat[]{%
      \label{fig:spcaps-100.3}
      \includegraphics[width=0.5\columnwidth]{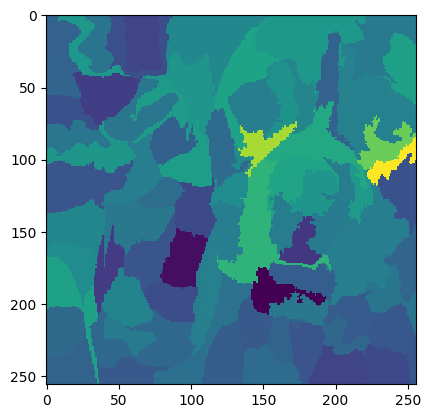}
    }
  \end{center}
  \caption{Supixelpixel-wise contribution to the entity feature vector visualized:
  on the left most column are input images;
  the rest of the columns (from left to right) are the superpixel contribution from Superpixel Capsule models with 10, 36 and 100 approximate superpixel numbers respectively.
  With relatively large superixel sizes, \textit{e.g.} Fig. \ref{fig:spcaps-16.3} and Fig. \ref{fig:spcaps-36.3} where the dog's eyes are captured by single superpixels,
  the models are conditioned to learn a vector representation of such parts of the the entity in question as superpixel features,
  allowing the later aggregation stages to construct the class features based on these superpixel features.
  Similar to the entity segmentation visualization from Fig. \ref{fig:sp-feature-space}, the superpixel feature contribution can provide very valuable information for segmenting the entity,
  where a higher response corresponds to a higher probability that the superpixel is part of the entity, according to Eq. (\ref{eq:spcaps-segmentations}).
  }
  \label{fig:spcaps-segmentations}
\end{figure*}

It is shown that the superpixel homogeneity assumption is not strictly necessary when superpixels are used to highlight various kinds of parts of the object in an image, as a means to part-whole analysis, as opposed to simply a dimensionality reduction technique.
From Fig. \ref{fig:compare_16} and Table \ref{tab:train}, one can learn that the use of very large superpixel sizes, 
where the image content within the superpixels are no longer just monotone texture (see also \ref{fig:spcaps-16.1} and \ref{fig:spcaps-36.1}, \textit{etc.}), 
can still produce a comparable predictive performance, as the curves for models \textit{spcaps-16} through \textit{spcaps-200} in Fig. \ref{fig:compare_16} show.
Here, the superpixel sizes are intentionally varied in order to allow the neurons to learn the superpixel features from different receptive field sizes, 
thereby changing the scale of the local context as the superpixel feature extraction reorganizes the feature space and readjusts the error back propagation accordingly during training.
As mentioned, the superpixel sizes are controlled indirectly by constraining the approximate number of superpixels per image upon superpixel segmentation.
Fig. \ref{fig:compare_16} shows the model convergence characteristics over 120 epoch of training by comparing the model configured to use 10, 16, 25, 36, 50, 100 and 200 superpixels per image sample respectively.
The general behavior that is observed is that with higher number of superpixels per image, the model appears to achieve slightly higher accuracy after 120 epochs of training 
and also exhibits slightly more overfitting.
However, as shown in Table \ref{tab:train}, all 7 models have generalized consistently well on this dataset and have a consistent validation accuracy at around 88\%.
%
In Fig. \ref{fig:compare_acc_16}, the gap that is seen between the model with 10 versus the higher number of superpixels can be attributed to the extreme violation of the superpixel homogeneity assumption by using very large superpixel sizes.
Overall, it has been confirmed that the superpixel sizes are an important hyperparameter for training this kind of part-whole recognition model, 
as they imply the scale of the neural receptive fields.

\begin{figure*}
  \begin{center}
    \subfloat[]{%
      \label{fig:spcaps-sizes-in}
      \includegraphics[width=0.5\columnwidth]{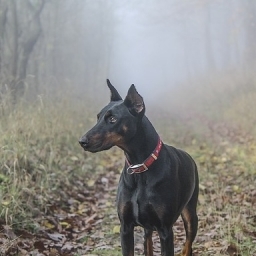}
    }
    \subfloat[]{%
      \label{fig:spcaps-sizes-10}
      \includegraphics[width=0.5\columnwidth]{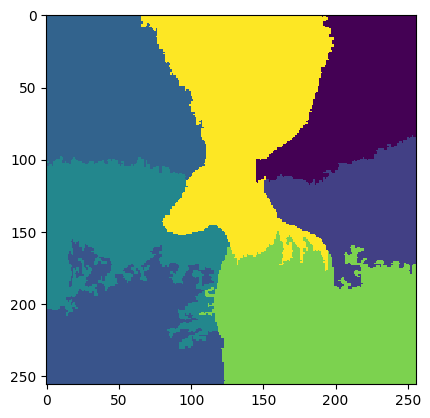}
    }
    \subfloat[]{%
      \label{fig:spcaps-sizes-16}
      \includegraphics[width=0.5\columnwidth]{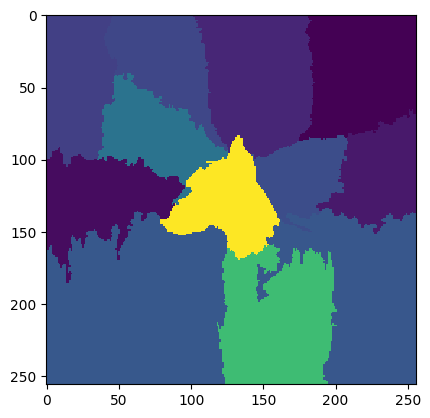}
    }
    \subfloat[]{%
      \label{fig:spcaps-sizes-25}
      \includegraphics[width=0.5\columnwidth]{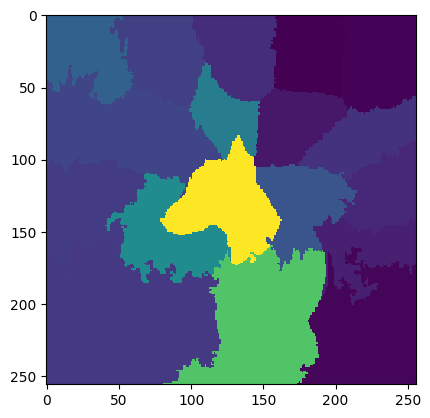}
    }
    
    \subfloat[]{%
      \label{fig:spcaps-sizes-36}
      \includegraphics[width=0.5\columnwidth]{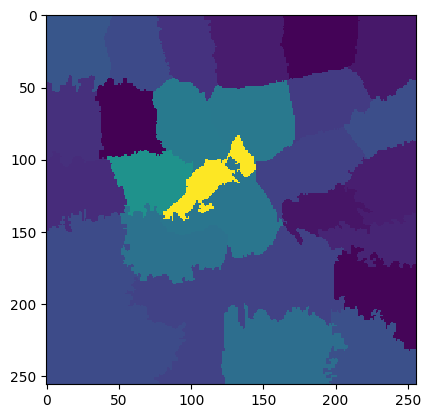}
    }
    \subfloat[]{%
      \label{fig:spcaps-sizes-50}
      \includegraphics[width=0.5\columnwidth]{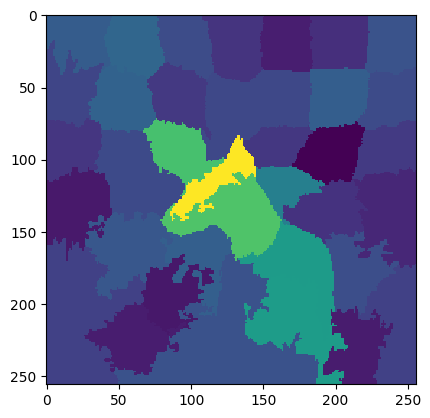}
    }
    \subfloat[]{%
      \label{fig:spcaps-sizes-100}
      \includegraphics[width=0.5\columnwidth]{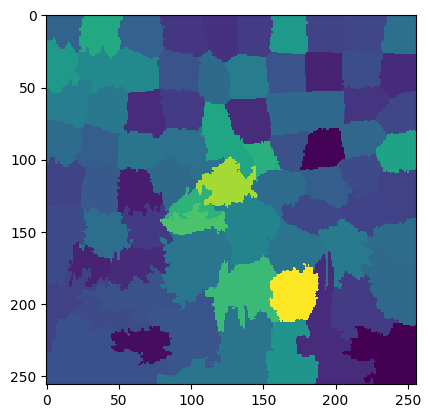}
    }
    \subfloat[]{%
      \label{fig:spcaps-sizes-200}
      \includegraphics[width=0.5\columnwidth]{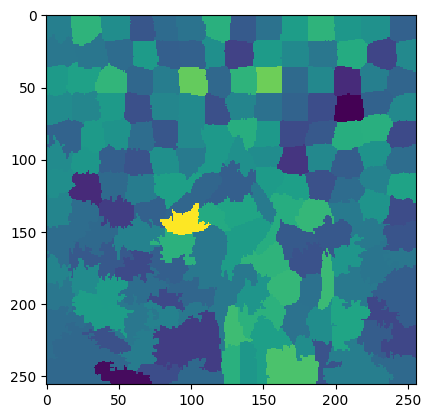}
    }

  \end{center}
  \caption{Supixelpixel-wise contribution to the class vector visualized with the sample sample across many different superpixel sizes.
  The approximated superpixels per image are constrained to be 10, 16, 25, 36, 50, 100, 200 to visualize what are the models learning in terms of part-whole relation modeling.
  An appropriate choice of superpixel size proved to be crucial to part-whole relation analysis, as shown in Fig. \ref{fig:spcaps-sizes-16} and Fig. \ref{fig:spcaps-sizes-25}.
  Counterexamples such as Fig. \ref{fig:spcaps-sizes-10}, has used a superpixel size too large, whereas Fig. \ref{fig:spcaps-sizes-50}, has created too many small superpixels, making the receptive field too small.}
  \label{fig:spcaps-segmentations-sizes}
\end{figure*}

\subsection{Visual explainability}\label{sect:sp-viz}

In this section, three sample images are drawn from three different classes (berry, bird and dog) to demonstrate the part-whole relation learning based on superpixels.
Another sample is presented with all the superpixel sizes that have been experimented with to provide a pictorial understanding of the training convergence characteristics from Fig. \ref{fig:compare_16} across different superpixel sizes.

In Fig. \ref{fig:spcaps-segmentations}, a sample of images were selected that have a relatively balanced mix of foreground and background in order to verify the part-whole relation expected to be learned by the model.
Fig. \ref{fig:spcaps-segmentations} visualizes the supixelpixel feature vector contribution to the class feature vector as a segmentation using a scalar projection described in Eq. (\ref{eq:spcaps-segmentations}).
These segmentations visualize the internal representation of the Superpixel Capsule Network that is learned without direct segment or parts training data, such as pixel level or superpixel level labels.

The models are conditioned to learn a vector representation of the superpixel features as relatively large superixel sizes were utilized,
where the superpixel homogeneity assumption would be clearly violated.
Depending on the scale of the parts and the entity in a given image,
an appropriate superpixel size can greatly assist the efficiency of such part-whole relation modeling.
For example, in Fig. \ref{fig:spcaps-36.1}, the superpixel sizes appear to coincide with the size of these blackberries, which in turn has facilitated an efficient representation of the superpixel feature for each backberry, which can eventually be transformed into a vector in class feature space per Eq. (\ref{eq:spcaps}).
On the contrary, in Fig. \ref{fig:spcaps-16.1}, the network had no choice but to come up with a superpixel feature representation for the group of blackberries and learn a way to map resulting superpixel feature vectors into the correct class feature vector in order to classify it correctly.
Intuitively, this would be much more difficult as the visual context that superpixel captures is so large that the same linguistic phrase ``a group of berries'' can have vastly different visual appearances.
Therefore, such representation would become less efficient,
which would explain why the ``spcaps-10'' curve in Fig. \ref{fig:compare_16} has shown a drastically slower convergence.
As a consequence of varying the superpixel sizes, 
the neurons responsible for learning the superpixel features will have different receptive field sizes.
By changing the scale of the local context as the superpixel feature extraction reorganizes the feature space and readjusting the error back propagation during training,
the models are learning a part-whole relationship subject to the constrains of the underlying superpixel segmentation, which provides the superpixel-wise neural receptive fields.

From their visual appearances, such as Fig. \ref{fig:spcaps-36.1}, Fig. \ref{fig:spcaps-16.2} and \ref{fig:spcaps-16.3}, one can confirm that the internal representation of part-whole relation in those models provides very useful information for semantic segmentation and object localization without requiring pixel-level or superpixel-level segmentation labels.

All of the segmentations have clear contours thanks to the underlying superpixel segmentation.
As the superpixel size decreases, or the number of superpixel increases, the strength of the responses tend to dilute and deteriorate. 
This can be observed in \textit{e.g.} Fig. \ref{fig:spcaps-100.2}, 
and it can be concluded that it is due to the fact that the superpixels and the entity capsules are fully connected, as described in Eq. (\ref{eq:spcaps}).
As each superpixel segmentation of the image is completely different, there is no specific role assigned to the neurons handling the superpixels and this makes the structure of the superpixel feature space more sporadic and difficult to learn with very high number of components.

This brings to curiosity the question of what the models are learning with much smaller superpixel sizes when the conventional superpixel homogeneity assumption is actually satisfied;
and therefore, another sample is presented (Fig. \ref{fig:spcaps-sizes-in}) along with the generated superpixel feature contribution visualizations of all the superpixel sizes.
As mentioned, an appropriate choice of superpixel size greatly assist an efficient part-whole analysis, as shown in Fig. \ref{fig:spcaps-sizes-16} and Fig. \ref{fig:spcaps-sizes-25}.
Fig. \ref{fig:spcaps-sizes-10} has clearly constrained the superpixel size to be too large to even correctly separate foreground and background, whereas Fig. \ref{fig:spcaps-sizes-50}, Fig. \ref{fig:spcaps-sizes-100} and Fig. \ref{fig:spcaps-sizes-200} has created too many small superpixels, making the receptive field too small for the models to come up with a good representation of the parts that comprise of the dog figure.

Additionally, it can be observed that the superpixel feature contribution from the models with very small superpixel sizes, \textit{e.g.} Fig. \ref{fig:spcaps-sizes-100} and Fig. \ref{fig:spcaps-sizes-200} are far less smooth and have more intensity oscillation across the 2-D spatial axes, which can be a sign of insufficient training due to the increase in model complexity.

\subsection{VGG-16 Classification Performance Baseline}

    

\begin{figure}
  \begin{center}
    \subfloat[]{%
      \label{fig:compare_acc_vgg16}
      \includegraphics[width=0.85\columnwidth]{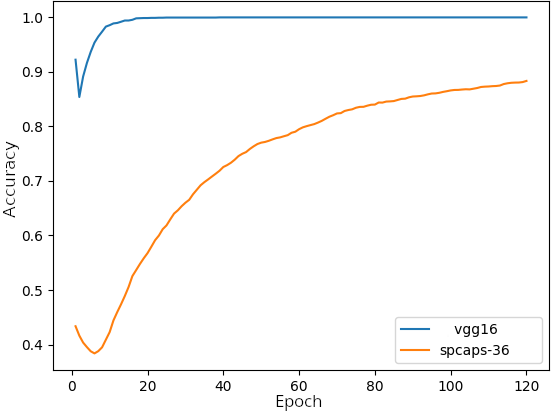}
      
    }
    
    \subfloat[]{%
      \label{fig:compare_loss_vgg16}
      \includegraphics[width=0.85\columnwidth]{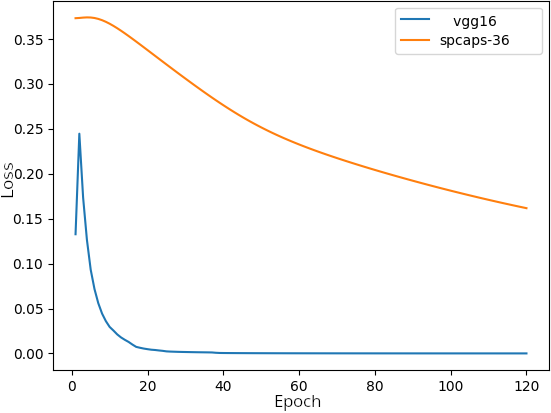}
    }
  \end{center}
  \caption{Comparison of training accuracy and loss between Superpixel Capsule Network and VGG-16 \cite{VGG}: both networks were initialized with VGG-16 pre-trained weights from ImageNet \cite{ILSVRC}.
  The standard VGG-16 network has achieved 96\% validation accuracy while the proposed network has reached 89\% validation accuracy.}
  \label{fig:compare_vgg16}
\end{figure}

All of the models contain the same convolutional layers from VGG-16 with the same pre-trained weights \cite{ILSVRC} as initialization.
Fig. \ref{fig:compare_vgg16} shows the differences during training between the Superpixel Capsule Network and a standard VGG-16 network, both using the same hyperparameter settings where possible.
The standard VGG-16 network has 134 million trainable parameters whereas the Superpixel Capsule has replaced the 120 million parameters from the VGG-16 dense layers with capsules containing approximately 3 million parameters.
The Superpixel Capsule Network used here is assuming approximately 36 superpixels per image.
With drastically fewer parameters compared to VGG-16, \textbf{17M vs 134M}, the model has achieved 89\% training accuracy.
In addition, that the validation accuracy has also reached 88\% at the same time with excellent generalization ability.
In these experiments, the 4-class dataset consisting of 6400 images is used with 25\% held out as validation data.
The initial oscillation in Fig. \ref{fig:compare_vgg16} is due to weight transferring where the neural networks had to adjust to the new target problem domain as well as the training data. 
This experiment has shown that the VGG-16 is still superior with regard to classification performance, but the proposed model has also achieved a satisfactory result with much fewer learnable parameters and a superior degree of explainability.
As shown in Table \ref{tab:train}, the VGG-16 has had slight overfitting based on the validation accuracy, but the proposed model has also achieved a more consistent inference results in the validation dataset as the training and validation accuracy are approximately the same.
However, the true strength with the proposed model is the ability to generate an entity segmentation by modeling the part-whole relationship, inherently trained with only classification labels (per Sect.~\ref{sect:sp-viz}).



      
    

\section{Conclusion and Future Directions}\label{sect:conclusion}

This paper has presented a novel neural architecture, \textit{Superpixel Capsule Network},
which leverages the heterogeneous superpixel in order to facilitate image object relational analysis.
The proposed model has demonstrated preliminary success in the effort of image object recognition based on relational analysis thanks to its capability to associate convolutional features to their spatial locality, and in turn, parts of objects in the image.
This architecture promotes explainability by allowing the inspection of the part-whole relationships within the image that factor into network's object recognition.
The ability to project the visual features and entity (capsule) activation back into the original superpixel segmentation offers a clear insight into what the model is learning and how it is aggregating said visual features into objects.
The resulting segmentation capability does not rely on segmetation labels as part of the training data, but only non-localized class labels.
From a segmentation model point of view, this fits better into the multiple-instance learning ideology, instead of direct segmentation training with labelled data.
In addition, the proposed model has drastically fewer learnable parameters compared with the existing state-of-the-art deep neural networks.
The new model has 88\% fewer of the total trainable parameters compared to VGG-16 and can achieve 89\% accuracy with very good generalization ability on the evaluation dataset.


Beyond the promising results here, there is still significant investigation needed to fully understand the behavior of this architecture.
For instance, to improve the model performance or to adapt to much larger datasets, research efforts will likely require more than simply using higher dimensional vector space.
This represents a future optimization challenge.
Future work will also extend the evaluation of the Superpixel Capsule Network, both in terms of broader dataset coverage and deeper analytical investigation of the architecture's behavior.

{\small
\bibliographystyle{unsrt} 
\bibliography{egbib}

\begin{thebibliography}{10}

\bibitem{ILSVRC}
J~Deng, A~Berg, S~Satheesh, H~Su, A~Khosla, and L~Fei-Fei.
\newblock Ilsvrc-2012, 2012.
\newblock {\em URL http://www. image-net. org/challenges/LSVRC}, 2012.

\bibitem{COCO}
Tsung-Yi Lin, Michael Maire, Serge Belongie, James Hays, Pietro Perona, Deva
  Ramanan, Piotr Doll{\'a}r, and C~Lawrence Zitnick.
\newblock Microsoft coco: Common objects in context.
\newblock In {\em European conference on computer vision}, pages 740--755.
  Springer, 2014.

\bibitem{FULLYCONV}
Jonathan Long, Evan Shelhamer, and Trevor Darrell.
\newblock Fully convolutional networks for semantic segmentation.
\newblock In {\em Proceedings of the IEEE conference on computer vision and
  pattern recognition}, pages 3431--3440, 2015.

\bibitem{SEMSEG}
Panqu Wang, Pengfei Chen, Ye~Yuan, Ding Liu, Zehua Huang, Xiaodi Hou, and
  Garrison Cottrell.
\newblock Understanding convolution for semantic segmentation.
\newblock In {\em 2018 IEEE Winter Conference on Applications of Computer
  Vision (WACV)}, pages 1451--1460. IEEE, 2018.

\bibitem{DCNF}
F.~Liu, C.~Shen, G.~Lin, and I.~Reid.
\newblock Learning depth from single monocular images using deep convolutional
  neural fields.
\newblock {\em IEEE Trans. Pattern Anal. Mach. Intell.}, 38(10), oct 2016.

\bibitem{DL3D}
Anastasia Ioannidou, Elisavet Chatzilari, Spiros Nikolopoulos, and Ioannis
  Kompatsiaris.
\newblock Deep learning advances in computer vision with 3d data: A survey.
\newblock {\em ACM Computing Surveys (CSUR)}, 50(2):20, 2017.

\bibitem{DEFORM}
Xizhou Zhu, Han Hu, Stephen Lin, and Jifeng Dai.
\newblock Deformable convnets v2: More deformable, better results.
\newblock In {\em Proceedings of the IEEE Conference on Computer Vision and
  Pattern Recognition}, pages 9308--9316, 2019.

\bibitem{SPSEG}
Mathijs Schuurmans, Maxim Berman, and Matthew~B Blaschko.
\newblock Efficient semantic image segmentation with superpixel pooling.
\newblock {\em arXiv preprint arXiv:1806.02705}, 2018.

\bibitem{SLIC}
R.~Achanta, A.~Shaji, K.~Smith, A.~Lucchi, P.~Fua, and S.~Susstrunk.
\newblock Slic superpixels compared to state-of-the-art superpixel methods.
\newblock {\em IEEE Trans. Pattern Anal. Mach. Intell.}, 34(11):2274–2282,
  nov 2012.

\bibitem{SPCNN}
Lifang Wu, Jiaoyu He, Meng Jian, Jianan Zhang, and Yunzhen Zou.
\newblock Fast cloud image segmentation with superpixel analysis based
  convolutional networks.
\newblock In {\em Systems, Signals and Image Processing (IWSSIP), 2017
  International Conference on}, pages 1--5. IEEE, 2017.

\bibitem{SPSEMSEG}
Mohammadreza Mostajabi, Payman Yadollahpour, and Gregory Shakhnarovich.
\newblock Feedforward semantic segmentation with zoom-out features.
\newblock In {\em Proceedings of the IEEE conference on computer vision and
  pattern recognition}, pages 3376--3385, 2015.

\bibitem{CAPS}
Sara Sabour, Nicholas Frosst, and Geoffrey~E Hinton.
\newblock Dynamic routing between capsules.
\newblock In {\em Advances in neural information processing systems}, pages
  3856--3866, 2017.

\bibitem{UNET}
Olaf Ronneberger, Philipp Fischer, and Thomas Brox.
\newblock U-net: Convolutional networks for biomedical image segmentation.
\newblock In {\em International Conference on Medical image computing and
  computer-assisted intervention}, pages 234--241. Springer, 2015.

\bibitem{pal2018capsdemm}
Anabik Pal, Akshay Chaturvedi, Utpal Garain, Aditi Chandra, Raghunath
  Chatterjee, and Swapan Senapati.
\newblock Capsdemm: Capsule network for detection of munro’s microabscess in
  skin biopsy images.
\newblock In {\em International Conference on Medical Image Computing and
  Computer-Assisted Intervention}, pages 389--397. Springer, 2018.

\bibitem{lalonde2018capsules}
Rodney LaLonde and Ulas Bagci.
\newblock Capsules for object segmentation.
\newblock {\em arXiv preprint arXiv:1804.04241}, 2018.

\bibitem{gSLIC}
Carl~Yuheng Ren, Victor~Adrian Prisacariu, and Ian~D Reid.
\newblock gslicr: Slic superpixels at over 250hz.
\newblock {\em arXiv preprint arXiv:1509.04232}, 2015.

\bibitem{EMCAPS}
Geoffrey~E Hinton, Sara Sabour, and Nicholas Frosst.
\newblock Matrix capsules with em routing.
\newblock {\em 6th International Conference on Learning Representations, ICLR},
  2018.

\bibitem{VGG}
K.~Simonyan and A.~Zisserman.
\newblock Very deep convolutional networks for large-scale image recognition.
\newblock In {\em Int. Conf. Learn. Representations}, 2015.

\bibitem{L5}
Giorgi Chaladze and Levan kalatozishvili.
\newblock Linnaeus 5 dataset for machine learning.
\newblock 2017.

\end{thebibliography}
}

\end{document}